\documentclass[10pt,twocolumn,letterpaper]{article}

\usepackage{cvpr}
\usepackage{times}
\usepackage{epsfig}
\usepackage{graphicx}
\usepackage{amsmath}
\usepackage{amssymb}

\usepackage{subfigure}
\usepackage{url}
\usepackage{bm}

\usepackage{multirow}

\usepackage[pagebackref=true,breaklinks=true,letterpaper=true,colorlinks,bookmarks=false]{hyperref}

\cvprfinalcopy 

\def\cvprPaperID{} 

\ifcvprfinal\fi
\begin{document}

\title{Precise Box Score: Extract More Information from Datasets to Improve the Performance of Face Detection}

\author{Ce Qi$^1$ \quad Xiaoping Chen$^1$ \quad Pingyu Wang$^1$ \quad Fei Su$^{1,2}$\\
$^1$School of Information and Communication Engineering\\
$^2$Beijing Key Laboratory of Network System and Network Culture\\
Beijing University of Posts and Telecommunications, Beijing, China\\
\\
{}
}

\maketitle

\begin{abstract}
For the training of face detection network based on R-CNN framework, anchors are assigned to be positive samples if intersection-over-unions (IoUs) with ground-truth are higher than the first threshold(such as 0.7); and to be negative samples if their IoUs are lower than the second threshold(such as 0.3). And the face detection model is trained by the above labels. However, anchors with IoU between first threshold and second threshold are not used. We propose a novel training strategy, Precise Box Score(PBS), to train object detection models. The proposed training strategy uses the anchors with IoUs between the first and second threshold, which can consistently improve the performance of face detection. Our proposed training strategy extracts more information from datasets, making better utilization of existing datasets. What's more, we also introduce a simple but effective model compression method(SEMCM), which can boost the performance of face detectors further. Experimental results show that the performance of face detection network can consistently be improved based on our proposed scheme.
\end{abstract}

\section{Introduction}

Face detection, which is the basis of face alignment and face recognition, plays an important role in face related tasks. More accurate face detection and face bounding boxes will also benefit the performance of face alignment and face recognition.

Many research works~\cite{zhu2012face,li2013probabilistic,li2014efficient,mathias2014face,chen2014joint,yang2014aggregate,li2015convolutional,yang2015facial,yang2015convolutional,zhu2017cms,jiang2017face,wang2017face,wang2017detecting,Hu_2017_CVPR,najibi2017ssh} have been done to improve the performance of face detectors. However, there is still big gap between humans and current face detectors, especially in the scenarios of small faces or occluded faces. The gap becomes bigger in case of resource constraint environment for the trading of the complexity and the required speed and memory. Some good performance face detectors are usually slow and high memory foot-prints(\emph{e.g.} it takes more than 1 second in \cite{Hu_2017_CVPR} per image and HyperNet~\cite{kong2016hypernet} is slow and has big model size).

One way to make the face detection models efficient is to use more powerful networks or design a specific network architecture. But this kind of strategy is not elegant and may cannot be used in other detection tasks. Recently, Hu \emph{et al.}~\cite{Hu_2017_CVPR} gets state-of-the-art results on the WIDER FACE detection benchmark~\cite{yang2016wider} by using a similar approach to the Region Proposal Networks(RPN)~\cite{ren2015faster} to directly detect faces. To boost the performance, it introduces an image pyramid as an integral part of the method, which is not that time efficient.

Another way is using more training data and data augmentation. AFLW~\cite{koestinger2011annotated}, which is a relatively smaller face detection dataset, is usually used as train set before. Face detection model's performance will be boosted when using the WIDER FACE dataset~\cite{yang2016wider}, a relatively bigger face detection dataset. What's more, data augmentation such as flipping and blurring will also help the final accuracy.

When training the R-CNN style face detector, anchors are assigned to be positive samples if intersection-over-unions (IoUs) with ground-truth are higher than the first threshold(such as 0.7); and to be negative samples if their IoUs are lower than the second threshold(such as 0.3). The object detection model is trained by the above labels, meaning the positive labels and negative labels are set by hand roughly. And the anchors with IoUs between first and second threshold are not used, which loses much information from detection dataset.

\begin{figure}
\centering
\subfigure[orignal R-CNN~\cite{ren2015faster} style training strategy]
{
\label{fig:orignal training strategy}
\includegraphics[width=0.22\textwidth]{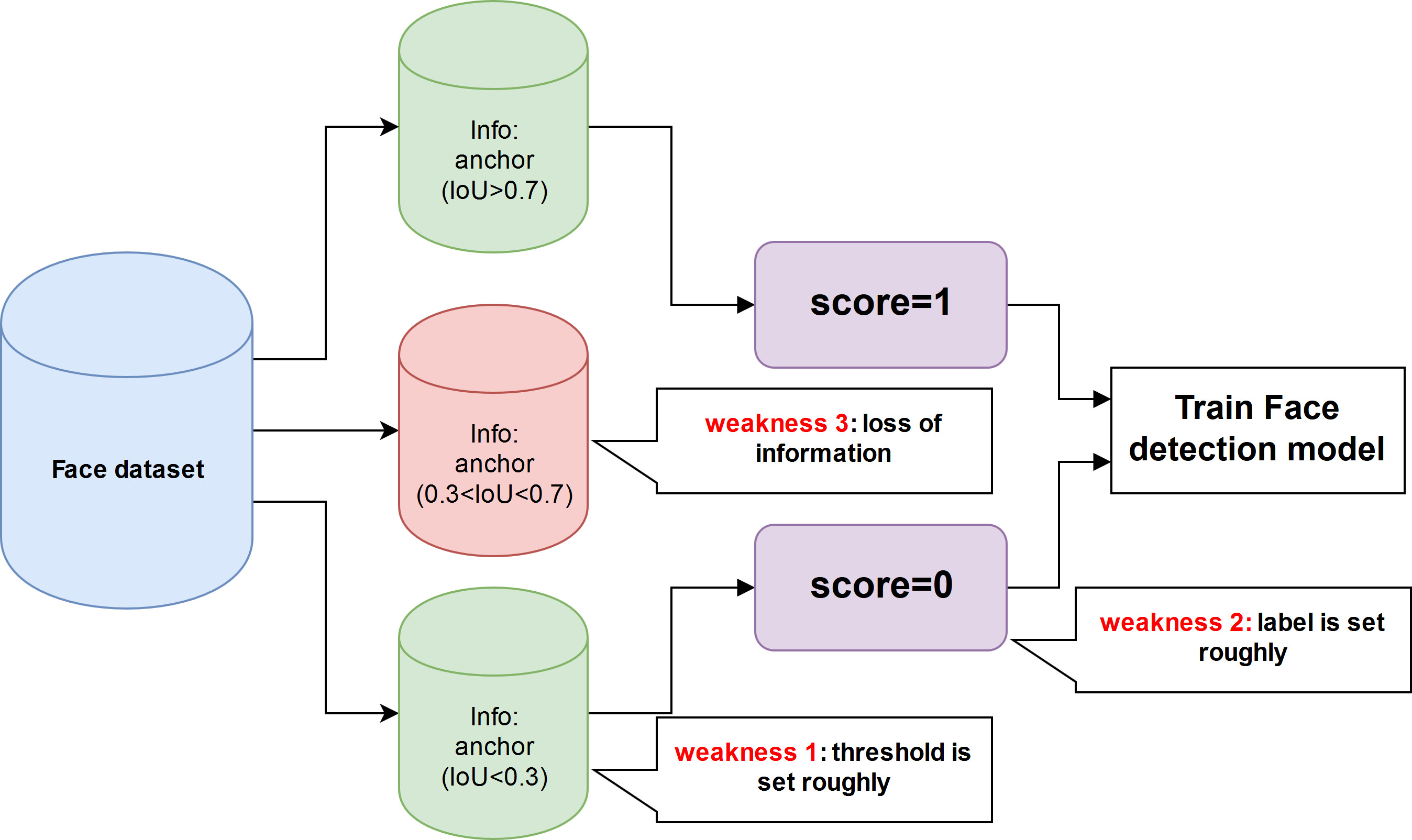}
}
\hspace{0in}
\subfigure[our new training strategy]
{
\label{fig:our new training strategy}
\includegraphics[width=0.22\textwidth]{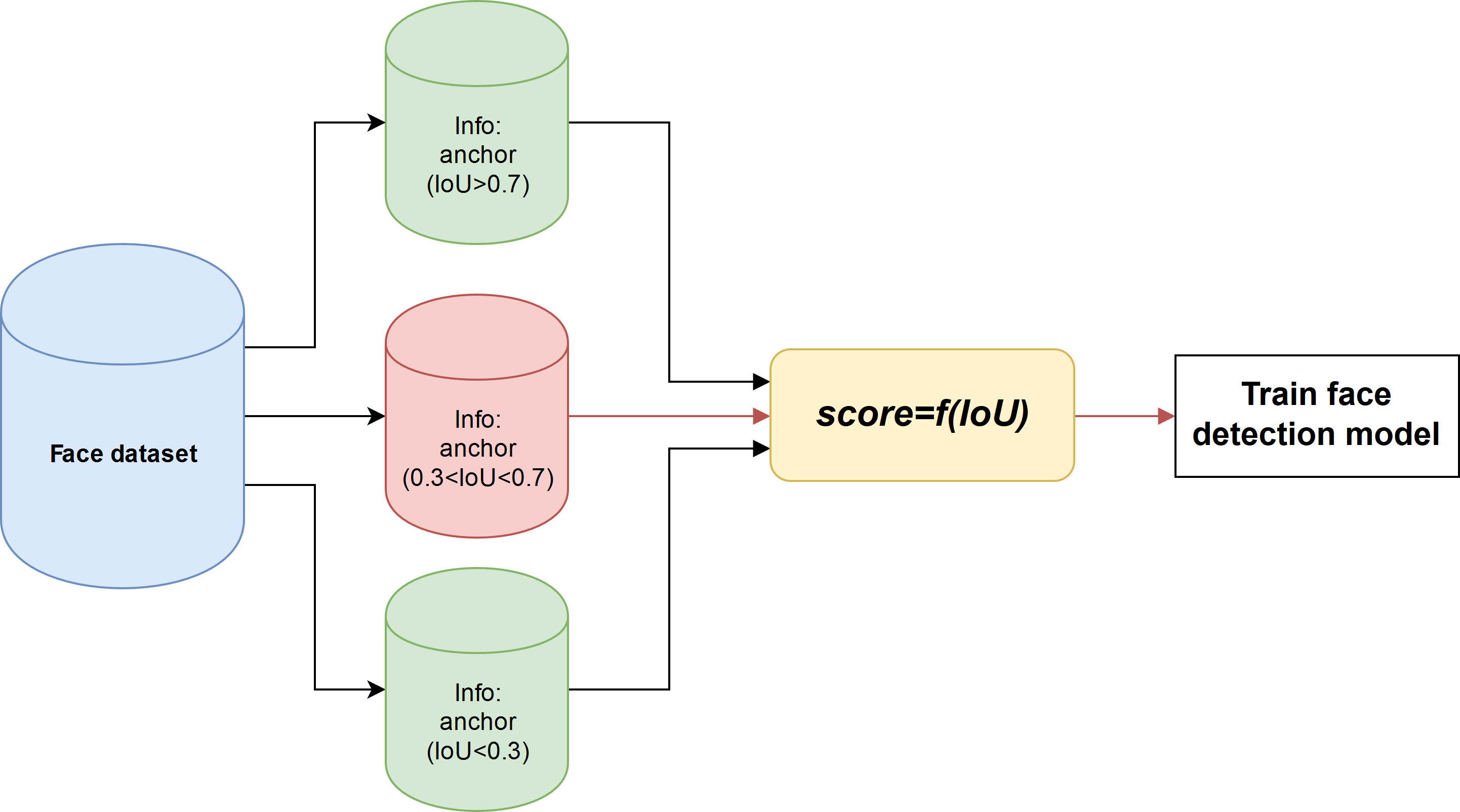}
}
\caption{Comparison of training strategy. Original R-CNN~\cite{ren2015faster} style training strategy uses incomplete information from face detection dataset(only use anchors with $IoU>0.7$ or $IoU<0.3$, and just set anchors $IoU>0.7$ with $score=1$ and $IoU<0.3$ with $score=0$ roughly). On the other hand, our new training strategy uses full information from face detection dataset and set the anchors's labels by a function precisely.}\label{fig:Comparison of training strategy}
\end{figure}

In this paper, we show that when training face detection model based on R-CNN framework, the original anchors assignment strategy is not appropriate and loses much information from original face detection dataset. As is shown in Fig. \ref{fig:orignal training strategy}, original training strategy has three weaknesses: (a) choosing thresholds roughly; (b) setting positive and negative labels with 1 or 0 roughly; (c) the information of anchors with IoUs between 0.7 and 0.3 is not used. So, we propose a novel training strategy, called Precise Box Score(PBS), to train face detection models. The proposed training strategy uses the anchors more effectively, meaning more information from face detection dataset will be used for training. What's more, we also introduce a simple but effective model compression method(SEMCM), which can boost the performance of face detectors further. The experimental results show that when using the proposed novel training strategy and model compression method, the performance of face detection model can consistently be improved.

The rest of the paper is organized as follows. Section \ref{sec:related work} provides an overview of the related works. Section \ref{sec:Novel Training strategy: Precise Box Score(PBS)} introduces the proposed training strategy: Precise Box Score(PBS) and the new architecture designed for PBS. Section \ref{sec:A Simple but Effective Model Compression Method(SEMCM)} describes a simple but effective model compression method(SEMCM), which can improve the performance of face detector and reduce the model size. Section \ref{sec:Experiments} presents the experiments and Section \ref{sec:Conclusion} gives the conclusion.

\section{Related Work}\label{sec:related work}

\subsection{Face Detection}
Face detection are the basis of face related tasks. And there are many works have been done to improve the performance of face detection. Before the re-emergence of convolutional neural networks(CNN), many traditional methods~\cite{viola2001rapid,zhu2012face,li2013probabilistic,li2014efficient,mathias2014face,chen2014joint,yang2014aggregate} have been proposed for face detection. The most successful traditional method is Viola-Jones~\cite{viola2001rapid} detector. However, most of the traditional methods use hand-crafted features, which limit the performance of face detector. Following the success of CNN~\cite{krizhevsky2012imagenet}, the performance of face detection is improved significantly, for the discriminative features of CNN. Recently, many CNN-based works have been done for face detection, such as~\cite{li2015convolutional,yang2015facial,yang2015convolutional,zhu2017cms,Hu_2017_CVPR}.

\subsection{R-CNN Style Face Detector}

The idea of detecting and localizing objects in two stages is widely used in object detection, such as Faster R-CNN~\cite{ren2015faster} and R-fcn~\cite{dai2016r}. Face detector~\cite{jiang2017face,wang2017face,wang2017detecting} with R-CNN style can obtain good accuracy. However, unlike the object detection with many classes, face detection detects only one class. Single stage face detectors~\cite{zhu2017cms,Hu_2017_CVPR,najibi2017ssh} also work well, which detect faces directly from the early convolutional layers with bounding box classification and regression. Most of the single stage face detection methods are more similar to the object proposal algorithm which is used as the first stage in detection pipeline. These kind of algorithms generally regress a set of anchors toward faces and assign scores to different anchors according to the intersection-over-unions (IoUs) between anchors and ground truth bounding boxes.

\subsection{Main Focus of Face Detection Research}

For single stage face detectors, there are many works have been done to boost the performance of face detection. Most of the methods focus on scale invariance and context modeling. Scale invariant can make detection of different scale of faces easier and context information can do help to hard classified faces.

For general object detection, ION~\cite{bell2016inside} uses skip pooling and RNN(recurrent neural networks) for context modeling and scale invariance. FPN~\cite{dollar2014fast} employs skip connections and multiple shared RPN from different convolutional layers. The same methods also be used for face detection. CMS-RCNN~\cite{zhu2017cms} employs skip connection, too. Hu~\etal~\cite{Hu_2017_CVPR} uses image pyramids and context modeling to improve the performance.

There are also some other methods focusing on object loss functions of detection, such as Unitbox~\cite{yu2016unitbox} and Grid loss~\cite{opitz2016grid}. Some other researchers do efforts on non-maximum suppression(NMS), a post processing step. Soft-NMS~\cite{bodla2017improving} uses a very simple but effective way to improve the NMS. Authors in \cite{hosang2016convnet} use a convolutional network to guide the NMS after detection.

Our training strategy(PBS) focuses on how to extract more information from current face detection dataset to improve the performance of face detection.

\section{Novel Training Strategy: Precise Box Score (PBS)}\label{sec:Novel Training strategy: Precise Box Score(PBS)}

Most of the works about detection are related to network architecture, loss function or post processing step. The main focus of the proposed training strategy is the input data step of model, just say, how to extract more information from current face detection datasets.

\begin{figure*}
\centering
\includegraphics[width=0.95\textwidth]{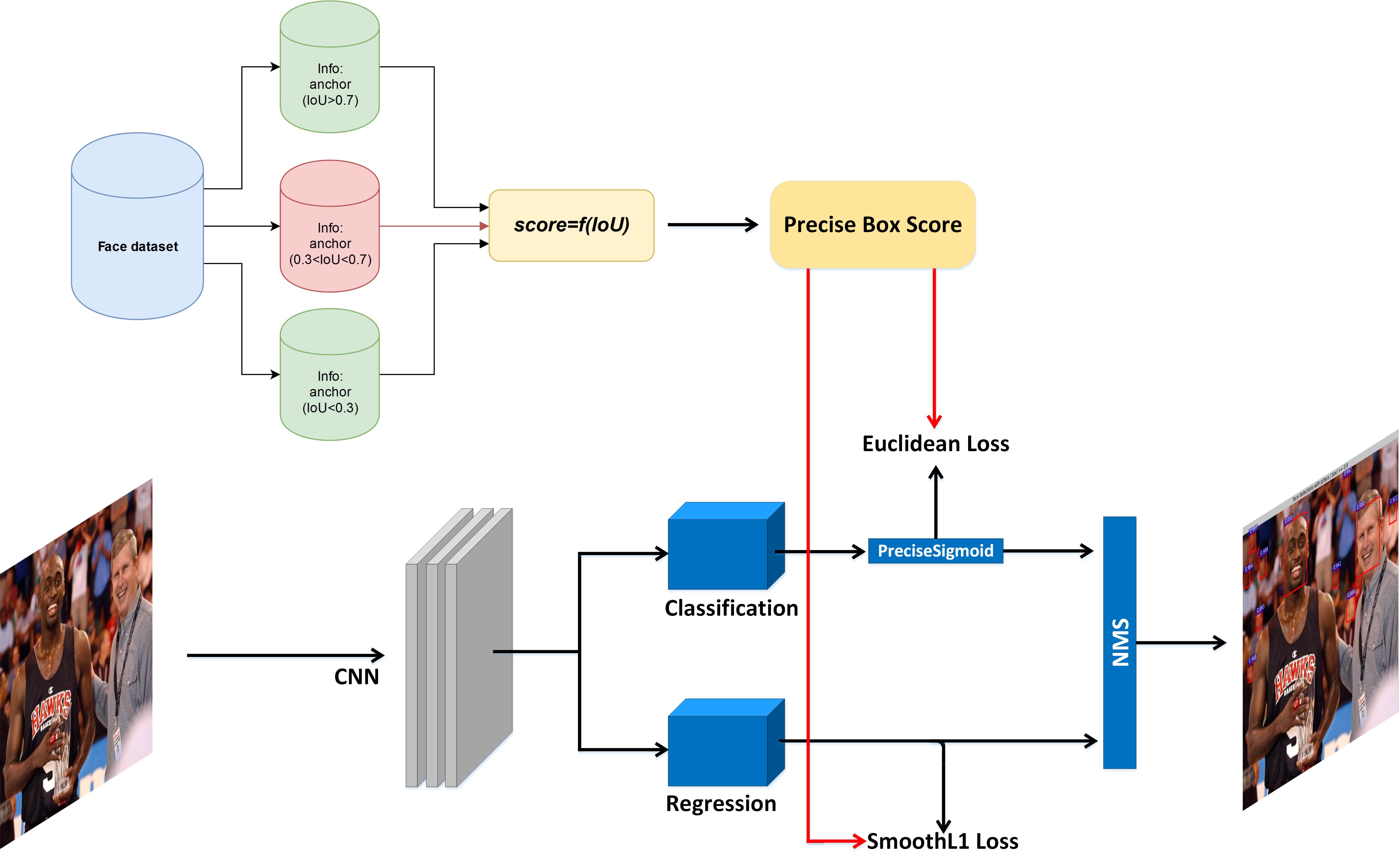}
\caption{Training and testing stages with PBS}\label{fig:PBS_train_test}
\end{figure*}

Our novel training strategy(PBS) is designed for detection network using anchors. Details are shown in Fig. \ref{fig:PBS_train_test}.

\subsection{General Architecture}

\begin{figure}
\centering
\subfigure[FDN with softmax]
{
\label{fig:FDN with softmax}
\includegraphics[width=0.45\textwidth]{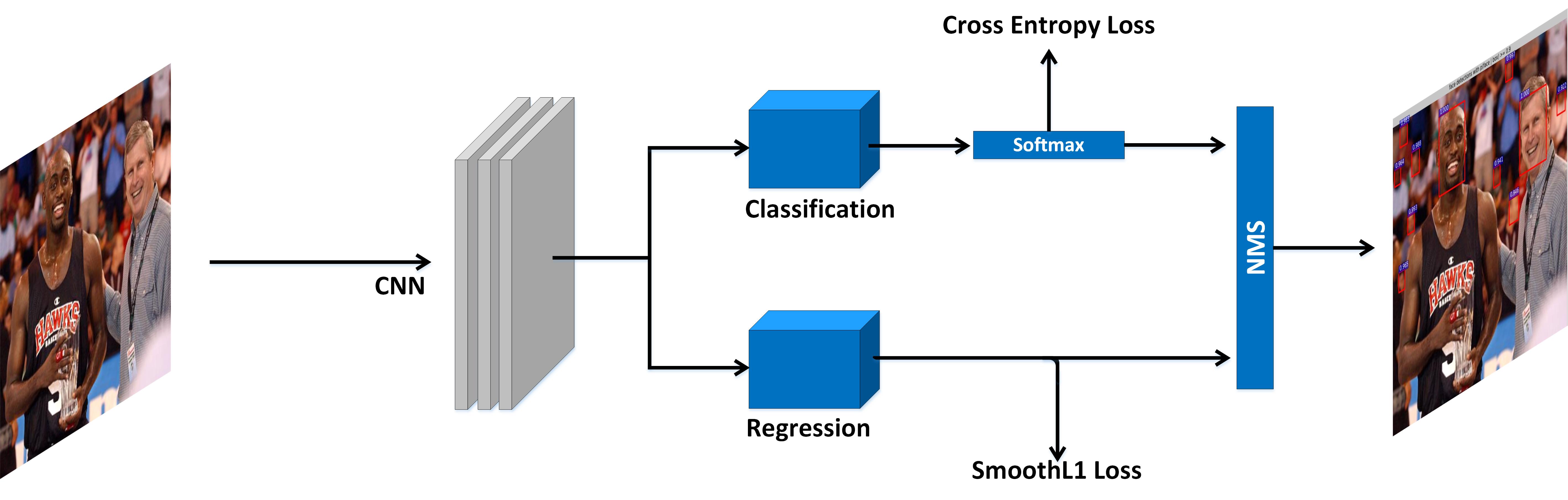}
}
\hspace{0in}
\subfigure[FDN with precise-sigmoid]
{
\label{fig:FDN with precise-sigmoid}
\includegraphics[width=0.45\textwidth]{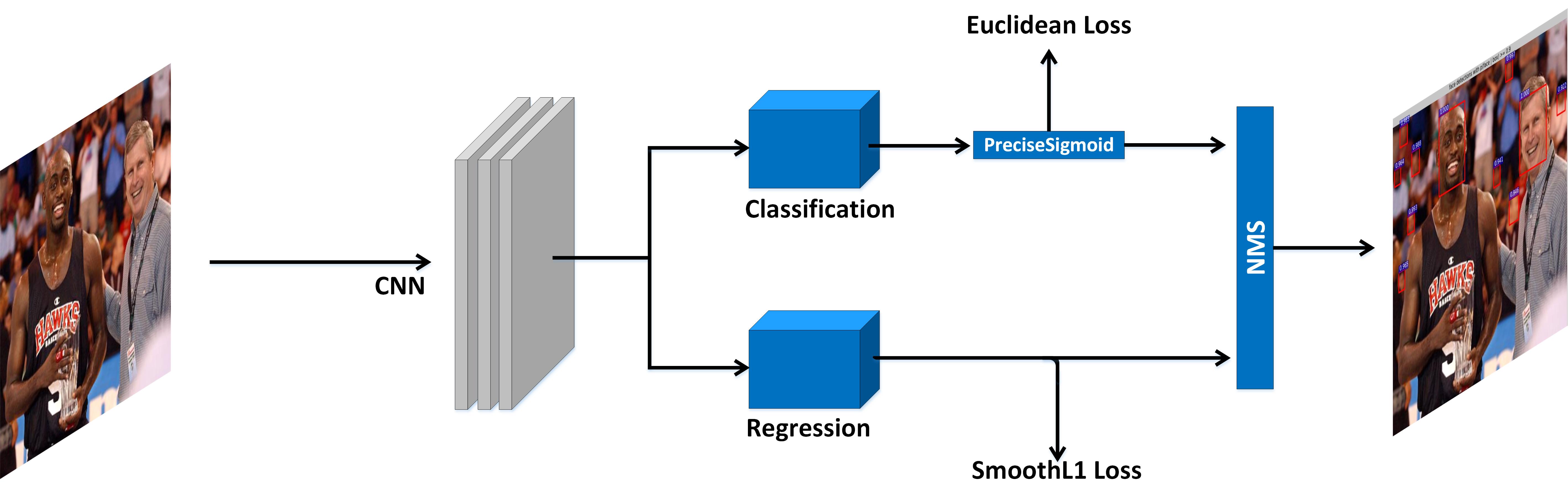}
}
\caption{General architecture of our face detection network(FDN). (a) is the network architecture of general one stage face detector using origianl training strategy as shown in Fig. \ref{fig:orignal training strategy}. (b) is a new network architecture proposed by ourself, designed for the proposed novel training stategy, Precise Box Score(PBS) as shown in Fig. \ref{fig:our new training strategy}.}\label{fig:General architecture of our face detection network(FDN)}
\end{figure}

Fig. \ref{fig:General architecture of our face detection network(FDN)} shows the general architecture we use, called Face Detection Network(FDN). It is a fully convolutional network which performs bounding box classification and regression simultaneously. For localization, just like RPN in \cite{ren2015faster}, our face detection network(FDN), regresses a set of predefined bounding boxes(anchors), to approximate the ground-truth bounding boxes. And the operations of bounding box classification and regression are added on the top of feature map with stride 16. The scales FDN used are 4, 8, 16 and 32. And, we only consider anchors with aspect ratio of 1:1 to reduce the number of total anchor boxes and fit the truth that most face boxes have aspect ratios of 1:1. Exactly, if the top of feature map with stride 16 has size $W_i \times H_i$, there would be $W_i \times H_i \times K $ anchors, where $K$ equals to $4$ in our setting.

For the reason of FDN with all convolutional network, the input images can be set to any size. And the total model size is small. Note that there are two different network architectures in Fig. \ref{fig:General architecture of our face detection network(FDN)}. Details will be described as follows.

\subsection{Precise Box Score(PBS)}\label{subsec:Precise Box Score(PBS)}

In this section, we will discuss the training strategy of R-CNN style detector, such as Faster R-CNN~\cite{ren2015faster}, R-fcn~\cite{dai2016r}, Hu \emph{et al.} ~\cite{Hu_2017_CVPR} and SSH~\cite{najibi2017ssh}. Furthermore, the details of our novel training strategy(PBS) will be given.

\textbf{General training strategy of R-CNN style detector:} For the training of R-CNN style detector, anchor boxes are introduced to serve as reference of multiple scales and aspect ratios. Classification and Regression are done simultaneously for anchors to do region proposal. When training, anchors are assigned a binary class label(of being an object or not). The conditions of assigning positive labels to anchors are: (i) anchors with the highest Intersection over Unions(IoUs) overlap with a ground-truth box, or (ii) anchors that have IoUs higher than $0.7$ with any ground-truth box. And anchors are assigned to be negative labels if their IoUs are lower than $0.3$ with all ground-truth boxes. Anchors that are neither positive nor negative do not contribute to the training. The detail of general training strategy of R-CNN style detector is shown in Fig. \ref{fig:orignal training strategy} and the formula is illustrated in Eq. (\ref{equ:orignal training strategy}). Note that the figure and formula above approximately illustrate the general training strategy of R-CNN style detector, for the existing of principle (i) described above.

\begin{equation}\label{equ:orignal training strategy}
label_{anchor}(IoU) =
\begin{cases}
$0$& \text{$IoU<0.3$}\\
$1$& \text{$IoU>0.7$}
\end{cases}
\end{equation}

\textbf{Weakness of general training strategy:} As shown in Fig. \ref{fig:orignal training strategy}, according to the general training strategy of R-CNN style described above, we can find that: (1) the positive anchors threshold($0.7$) and negative anchors threshold($0.3$) are set to certain numbers roughly. The two thresholds in Faster R-CNN are $0.7$ and $0.3$ respectively. And the two thresholds in R-fcn are the same numbers, $0.5$. (2) the labels are binary class, meaning the labels are $0$ or $1$. The rough binary label loses much information from detection dataset. For example, the anchor with IoU $0.75$ is different with the anchor with IoU $1.0$. The latter is more like a positive sample than the former. (3) the anchors with IoUs between first and second thresholds are not used.

\textbf{Precise Box Score(PBS):} To overcome the three weaknesses of general training strategy described above, we propose a novel training strategy, Precise Box Score(PBS). PBS will choose the best thresholds and use precise float point numbers as labels, when the precise float point numbers are the outputs of a designed function using IoUs as inputs, as illustrated in Fig. \ref{fig:our new training strategy}. It will firstly choose the best thresholds through experiments and then choose a best function to translate IoUs to labels. Detailed steps are as follows:

(1) Using Eq. (\ref{equ:pbs_just_adjust_pos_bound}) to choose the best thresholds through experiments(Note that $Bound_{pos} = 0.3, 0.4, 0.5, 0.7$ when $Bound_{neg}=0.3$, and $Bound_{pos}=0.2$ when $Bound_{neg}=0.1$). The best thresholds are represented by $Bound_{best\_pos}$ and $Bound_{best\_neg}$.

(2) Choosing the best function to translate IoUs to labels, based on best thresholds $Bound_{best\_pos}$ and $Bound_{best\_neg}$ obtained in step (1). Three classes of functions are used, as illustrated in Eq. (\ref{equ:pbs_add}), Eq. (\ref{equ:pbs_split_one}) and Eq. (\ref{equ:pbs_split_two}). Eq. (\ref{equ:pbs_add}) adds a shift variable A($A \ge 0$) to the IoU for the positive label, and when the positive label is bigger than $1$, it will be set to $1$. This function roughly translates the IoUs to labels. Eq. (\ref{equ:pbs_split_one}) limits some IoUs($Bound_{best\_pos} \textless IoU \textless Bound_1$) to $Score_1$. And Eq. (\ref{equ:pbs_split_two}) uses more variables and do more precise limitation to the function of IoU. Details will be shown in experiments in Section \ref{subsec:Precise Sigmoid+PBS on FDDB}.

\begin{equation}\label{equ:pbs_just_adjust_pos_bound}
label_{anchor}(IoU) =
\begin{cases}
$0$& \text{$IoU \textless Bound_{neg}$}\\
$1$& \text{$IoU \textgreater Bound_{pos}$}
\end{cases}
\end{equation}

\begin{equation}\label{equ:pbs_add}
\resizebox{0.9\hsize}{!}
{$
label_{anchor}(IoU) =
\begin{cases}
$0$& \text{$IoU \textless Bound_{best\_neg}$}\\
$IoU+A$& \text{IoU \textgreater $Bound_{best\_pos}, (IoU+A) \textless 1$}\\
$1$& \text{$IoU \textgreater Bound_{best\_pos}, (IoU+A) \ge 1$}
\end{cases}
$}
\end{equation}

\begin{equation}\label{equ:pbs_split_one}
\resizebox{0.9\hsize}{!}
{$
label_{anchor}(IoU) =
\begin{cases}
$0$& \text{$IoU \textless Bound_{best\_neg}$}\\
$$Score_1$$& \text{$Bound_{best\_pos} \textless IoU \textless Bound_1$}\\
$1$& \text{$IoU \ge Bound_1$}
\end{cases}
$}
\end{equation}

\begin{equation}\label{equ:pbs_split_two}
\resizebox{0.9\hsize}{!}
{$
label_{anchor}(IoU) =
\begin{cases}
$0$& \text{$IoU \textless Bound_{best\_neg}$}\\
$$Score_1$$& \text{$Bound_{best\_pos} \textless IoU \textless Bound_1$}\\
$$Score_2$$& \text{$Bound_1 \le IoU \textless Bound_2$}\\
$1$& \text{$IoU \ge Bound_2$}
\end{cases}
$}
\end{equation}

\subsection{New Architecture Designed for Precise Box Score(PBS)}

As introduced above, the proposed novel training strategy, Precise Box Score(PBS), uses precise float point numbers as labels, not simply uses binary labels. So, softmax with cross entropy loss, as shown in Eq. (\ref{equ:softmax with cross entropy loss}), which is designed for binary labels, is not appropriate for the precise float point number labels.


\begin{equation}\label{equ:softmax with cross entropy loss}
L_{s}=-\frac{1}{N}\sum_{i} logP(y_i|x_i)=-\frac{1}{N}\sum_{i} log\frac{e^{f_{y_i}}}{\sum_{j} e^{f_j}}
\end{equation}


\noindent where $x_i$ and $y_i \in [1...C]$ denote the $ith$ input data and its corresponding label, respectively. $f_j$ denotes the $jth$ element of the softmax input vector $f$, and $j \in [1...C]$. $N$ is the number of training images. $C$ is the number of class.

We design a new architecture designed for Precise Box Score(PBS), as illustrated in Fig. \ref{fig:FDN with precise-sigmoid}. Because of two reasons, the new architecture is needed:

(a) Our proposed novel training strategy, Precise Box Score(PBS), uses precise float point numbers as labels, not simply uses binary labels.

(b) The precise float point number labels the PBS used are in $[0,1]$.

So, the designed new architecture for PBS replaces the softmax(with cross entropy loss)with sigmoid(with euclidean loss), as shown in Eq. (\ref{equ:sigmoid with euclidean loss}), which is called Precise Sigmoid. The new architecture, which uses sigmoid(with euclidean loss), will output the numbers in $[0,1]$, while with the loss for precise float point number.

The formula of sigmoid with euclidean loss(Precise Sigmoid) is shown in Eq. (\ref{equ:sigmoid with euclidean loss}).

\begin{equation}\label{equ:sigmoid with euclidean loss}
L_{PreciseSigmoid}=\frac{1}{2 \times N}\sum_{i} {(\frac{1}{1+e^{-x_i}}-y_i)}^2
\end{equation}


\noindent where $L_{PreciseSigmoid}$ denotes the Precise Sigmoid Loss, new loss designed for PBS in the new architecture. $x_i$ and $y_i \in [1...C]$ denote the $ith$ input data and its corresponding label. $N$ is the number of training images. $C$ is the number of class.

The loss of our face detector using new architecture is:

\begin{equation}\label{equ:new final loss}
L=\lambda  L_{PreciseSigmoid} + L_{Regression}
\end{equation}

\noindent where $L_{PreciseSigmoid}$. denotes the Precise Sigmoid Loss. $L_{Regression}$ denotes the SmoothL1 Loss~\cite{ren2015faster} used for bounding box regression.

\textbf{Note:} we know that in mathematics, the sigmoid with euclidean loss may generate gradient vanishing. To solve this problem, we train the new architecture based on the parameters of a model pretrained by softmax with cross entropy loss. Experiments will be given in Section \ref{sec:Experiments} to demonstrate the effectiveness of this method, which can avoid the gradient vanishing of sigmoid with euclidean loss. The experimental results for demonstration are shown in Table \ref{table:accuracy comparison of precise sigmoid and softmax}.

The benefits of new architecture are as follows:

(a) Training phase: using precise float point numbers in $[0,1]$ as labels satisfies the training request of PBS.

(b) Testing phase: the outputs of sigmoid are in $[0,1]$, which can be used as scores of face boxes directly.

(c) It is effective to reduce the params of models through sigmoid, which makes the training and testing faster.

\subsection{Superiority of Precise Sigmoid+Precise Box Score(PBS)}

(a) Using the labels of PBS with Precise Sigmoid is the full implementation of proposed new training strategy.

(b) Using the precise float point number labels can extract more information from detection dataset, which can help the training.

(c) Under PBS, models can output precise and appropriate scores for bounding boxes, which can benefit the post processing of NMS(bounding boxes with lower IoUs with ground-truth get lower scores).

\section{A Simple but Effective Model Compression Method(SEMCM)}\label{sec:A Simple but Effective Model Compression Method(SEMCM)}

\begin{figure*}
\centering
\includegraphics[width=0.95\textwidth]{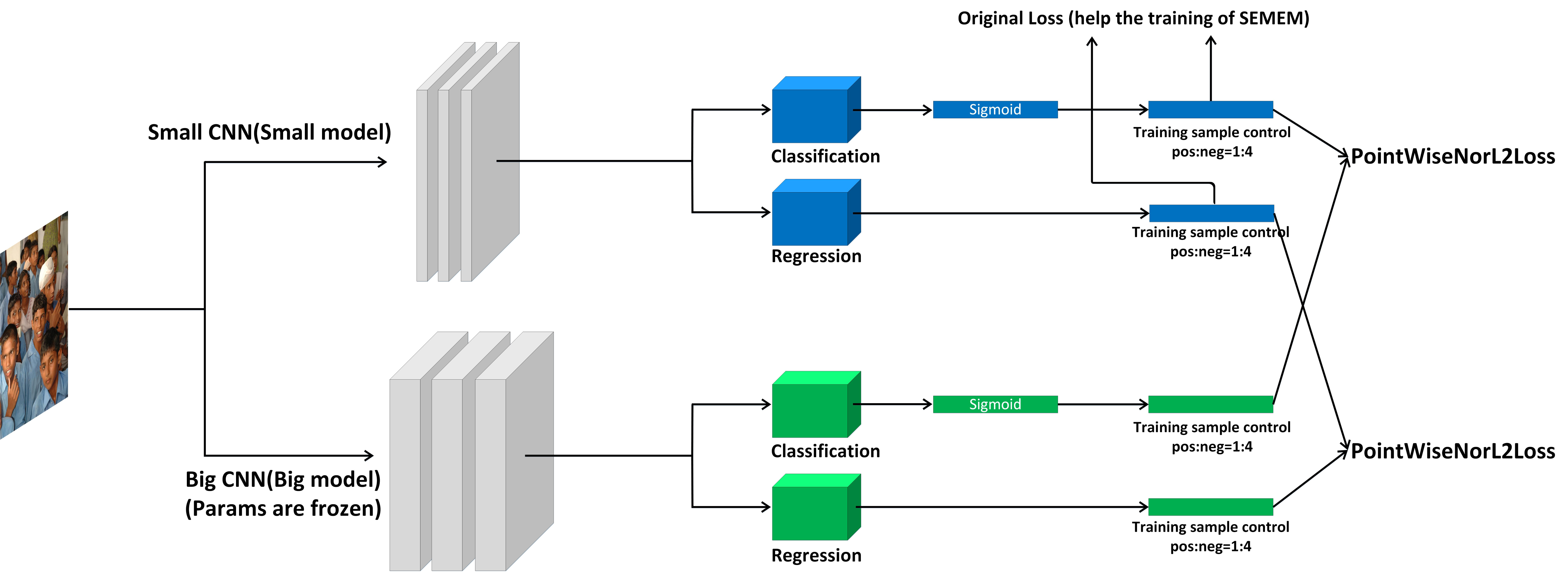}
\caption{Overall architecture of SEMCM.}\label{fig:overall architecture of SEMCM}
\end{figure*}

After using the Precise Box Score(PBS) to improve the performance of face detection model, we propose a simple but effective model compression method(SEMCM) for one stage face detector. There are also some other works for model compression~\cite{han2015deep,hinton2015distilling,luo2016face,li2017mimicking}. ~\cite{han2015deep} is a general but relatively complex model compression method. \cite{hinton2015distilling,luo2016face,li2017mimicking} are relatively simple model compression methods, while ~\cite{hinton2015distilling,luo2016face} are designed for image classification or face recognition. \cite{li2017mimicking} is designed for object detection, but this method is used for two stage detector and uses the information of Region Of Interest(ROI), which does not exist in one stage detector.

SEMCM is designed for one stage detector, as illustrated in Fig. \ref{fig:overall architecture of SEMCM}. One stage detector uses classification and regression results for anchors simultaneously to get the final results. SEMCM uses the output from pretrianed big model directly and the output will be used as supervision signals for the training of small model. Note that the output feature maps of big model and small model should be in same dimension in width, height and channel. So, the most convenient way to get a trainable small model is just downsampling the channels of all layers in big model, except the output layers used for classification and regression. The details of training steps of SEMCM are described as follows:

(1) A big model is trained through our proposed training strategy(PBS).

(2) A small model is obtained by downsampling the channels of all layers in big model, except the output layers used for classification and regression.

(3) The small model is trained simply by general training strategy with some iterations. SEMCM also works for half-trained small model.

(4) Frozen the parameters of pretrained big model in (1), and train the small model supervised by the output feature maps of big model.

(5) \textbf{Note:} to make the training of SEMCM stable. Two new layers are designed, to guide how to use the output feature maps of big model to supervise the training of small model, as illustrated in Fig. \ref{fig:overall architecture of SEMCM}. And the original training signals are also added to guarantee the good performance of SEMCM.

\section{Experiments}\label{sec:Experiments}

\subsection{Experimental Setup}

The new architecture models(shown in Fig. \ref{fig:FDN with precise-sigmoid}) trained with PBS will start the training from a pretrained network trained through softmax with cross entropy loss. Using a pretrained model will solve the problem of gradient vanishing of sigmoid with euclidean loss for PBS. All anchors have aspect ratio of 1:1. For FDDB~\cite{fddbTech}, anchors with scales \{4, 8, 16, 32\} are used on the feature map with total stride $16$. The training and testing are both single scale, meaning we rescale the shorter side of the image up to 600 pixels while keeping the longer side below 1000 pixels without changing the aspect ratio. For WIDER FACE dataset~\cite{yang2016wider}, all settings follow the SSH~\cite{najibi2017ssh}. During inference, model outputs 300 top scoring boxes and NMS with threshold of $0.3$ is performed on the boxes to get the final detection results.

The goal of our experiments is to demonstrate the effectiveness of PBS and SEMCM, so we use relatively simple networks to verify the effectiveness of our methods. We use one stage face detector with main bone of ZF-net~\cite{zeiler2014visualizing}, VGG\_CNN\_M\_1024~\cite{chatfield2014return}, VGG16~\cite{simonyan2014very} and ZF-24-net. Note that the ZF-24-net is a network designed for SEMCM. ZF-24-net has the same architecture as ZF-net, except all layers' channel reduced by $\frac{1}{4}$(not inculde layers for classification and regression), meaning ZF-24-net is a smaller network compared with ZF-net. The four designed one stage face detectors' model size is 17.3M, 30.9M, 68.3M and 1.1M, respectively.

\subsection{Datasets}

FDDB~\cite{fddbTech} and WIDER FACE~\cite{yang2016wider} are used in our experiments.

\textbf{FDDB~\cite{fddbTech}:} FDDB contains 2845 images with 5171 annotated faces. We use this dataset only for testing.

\textbf{WIDER FACE~\cite{yang2016wider}:} WIDER FACE contains 32, 203 images with 393, 703 annotated faces, 158, 989 of which are in the train set, 39, 496 in the validation set and rest are in the test set. The validation and test set are divided into ``easy'', ``medium'', ``hard'' subsets cumulatively(\emph{i.e.} the ``hard'' set contains all images). This is one of the most challenging public face detection datasets, with wide variety of face scales and occlusion. By default, we train our models on the train set of WIDER FACE~\cite{yang2016wider} and evaluate on the validation set of WIDER FACE or FDDB~\cite{fddbTech}.

\subsection{Ablation study of loss weight for Precise Sigmoid}

We firstly do experiments on FDDB~\cite{fddbTech} for loss weight $\lambda$ in Eq. (\ref{equ:new final loss}) of the proposed new architecture(shown in Fig. \ref{fig:FDN with precise-sigmoid}) for PBS. When doing the experiments for loss weight of $L_{PreciseSigmoid}$ in Eq. (\ref{equ:new final loss}), we use the original R-CNN style training strategy, as described in Section \ref{subsec:Precise Box Score(PBS)}, and the two thresholds are set to $0.7$ and $0.3$. All anchors have aspect ratio of 1:1 with scales \{4, 8, 16, 32\}, which are used on the feature map with total stride $16$.

We use the train set of WIDER FACE for training and FDDB for testing. Different networks are used as main bone of one stage detector, to find an optimal loss weight for different networks. The results are shown as follows. Table \ref{table:accuracy on FDDB with different loss weight of zf-net},\ref{table:accuracy on FDDB with different loss weight of vgg_m_cnn_1024} and \ref{table:accuracy on FDDB with different loss weight of zf_24} give the results of one stage detector with main bone of ZF-net~\cite{zeiler2014visualizing}, VGG\_CNN\_M\_1024~\cite{chatfield2014return} and ZF-24-net. the accuracy is measured when the false positive is $1000$, $500$, $100$ respectively on FDDB. The ``$-$'' in three tables means the model cannot converge well.

\begin{table}[]
\centering
\scalebox{0.65}
{
\begin{tabular}{|c|c|c|c|}
\hline
\multirow{2}{*}{Loss weight($\lambda$)} & \multicolumn{3}{c|}{Accuracy(\%)} \\ \cline{2-4}
                             & 1000 FP    & 500 FP    & 100 FP    \\ \hline
1&-&-&- \\ \hline
10&85.7&79.4&51.4 \\ \hline
20&86.6&81.1&61.3 \\ \hline
100&91.1&87.9&74.2 \\ \hline
200&92.4&90.5&81.4 \\ \hline
\textbf{300}&\textbf{92.7}&\textbf{90.7}&\textbf{83.2} \\ \hline
400&92.6&91.0&82.5 \\ \hline
\end{tabular}
}
\vspace{1em}
\caption{Accuracy(\%) on FDDB with different loss weight $\lambda$ for $L_{PreciseSigmoid}$. The main bone of the detector is \textbf{ZF-net}.}
\label{table:accuracy on FDDB with different loss weight of zf-net}
\end{table}

\begin{table}[]
\centering
\scalebox{0.65}
{
\begin{tabular}{|c|c|c|c|}
\hline
\multirow{2}{*}{Loss weight($\lambda$)} & \multicolumn{3}{c|}{Accuracy(\%)} \\ \cline{2-4}
                             & 1000 FP    & 500 FP    & 100 FP    \\ \hline
1&-&-&- \\ \hline
10&89.5&87.6&72.8 \\ \hline
20&89.6&86.8&72.0 \\ \hline
100&91.6&89.9&79.8 \\ \hline
200&91.8& 90.0&80.2 \\ \hline
\textbf{300}&\textbf{92.2}&\textbf{90.5}&\textbf{82.3} \\ \hline
400&92.1&90.4&81.1 \\ \hline
\end{tabular}
}
\vspace{1em}
\caption{Accuracy(\%) on FDDB with different loss weight $\lambda$ for $L_{PreciseSigmoid}$. The main bone of the detector is \textbf{VGG\_CNN\_M\_1024}.}
\label{table:accuracy on FDDB with different loss weight of vgg_m_cnn_1024}
\end{table}

\begin{table}[]
\centering
\scalebox{0.65}
{
\begin{tabular}{|c|c|c|c|}
\hline
\multirow{2}{*}{Loss weight($\lambda$)} & \multicolumn{3}{c|}{Accuracy(\%)} \\ \cline{2-4}
                             & 1000 FP    & 500 FP    & 100 FP    \\ \hline
1&-&-&- \\ \hline
100&88.6& 86.9&78.6 \\ \hline
200&89.3&87.4&80.9 \\ \hline
\textbf{300}&\textbf{89.8}&\textbf{88.2}&\textbf{82.3} \\ \hline
400&89.8&88.1&82.3 \\ \hline
500&89.8&88.1&82.2 \\ \hline
\end{tabular}
}
\vspace{1em}
\caption{Accuracy(\%) on FDDB with different loss weight $\lambda$ for $L_{PreciseSigmoid}$. The main bone of the detector is \textbf{ZF-24-net}.}
\label{table:accuracy on FDDB with different loss weight of zf_24}
\end{table}

From Table \ref{table:accuracy on FDDB with different loss weight of zf-net},\ref{table:accuracy on FDDB with different loss weight of vgg_m_cnn_1024} and \ref{table:accuracy on FDDB with different loss weight of zf_24}, the optimal loss weight $\lambda$ equals to $300$. So, for the new architecture with Precise Sigmoid, we use $\lambda=300$ for $L_{PreciseSigmoid}$ by default.

\subsection{Precise Sigmoid+PBS on FDDB}\label{subsec:Precise Sigmoid+PBS on FDDB}

\textbf{The comparable performance of Precise Sigmoid}: we also compare the performance of our Precise Sigmoid with softmax \textbf{by using original R-CNN style training strategy}, as shown in Table \ref{table:accuracy comparison of precise sigmoid and softmax}. The results show the comparable performance of Precise Sigmoid and softmax. Furthermore, better results on VGG16 based network of Precise Sigmoid prove the gradient vanishing problem can be solved.

\textbf{The effectiveness of Precise Sigmoid+PBS:} the Precise Sigmoid is designed for PBS, and experimental results will show the effectiveness of Precise Sigmoid+PBS. The experiments are done with ZF-net as main bone of face detector.

\begin{figure}
\centering
\includegraphics[width=0.46\textwidth]{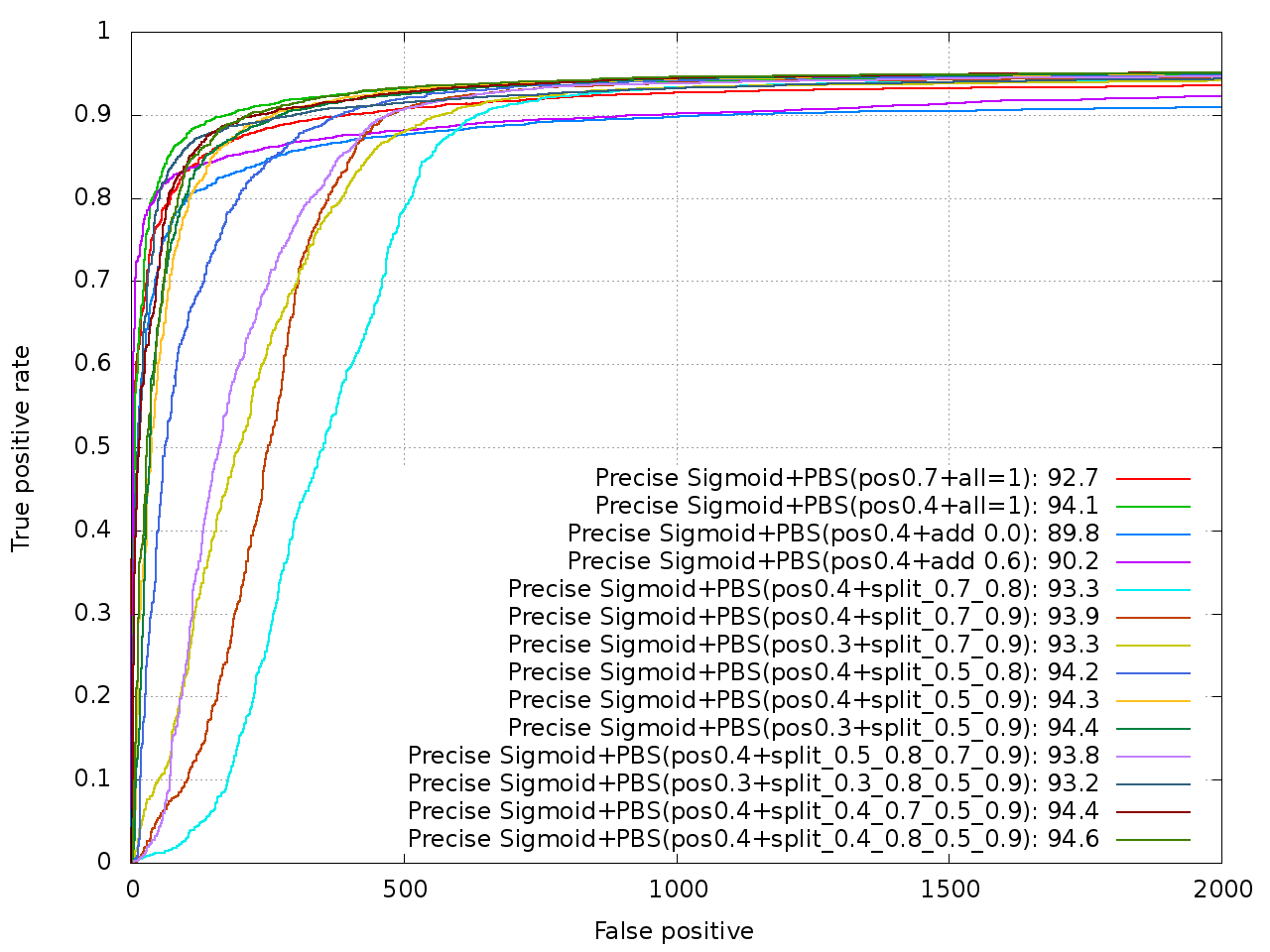}
\caption{Detailed experimental results of different Precise Box Score(PBS) on FDDB. ``pos0.7+all=1'' denotes the using of Eq. (\ref{equ:pbs_just_adjust_pos_bound}) with $Bound_{pos}=0.7$. ``pos0.4+add0.6'' denotes the using of Eq. (\ref{equ:pbs_add}) with $Bound_{pos}=0.4$ and $A=0.6$. ``pos0.4+split\_0.7\_0.8'' demotes the using of Eq. (\ref{equ:pbs_split_one}) with $Bound_{best\_pos}=0.4$, $Bound_1=0.7$ and $Score_1=0.8$. ``pos0.4+split\_0.4\_0.8\_0.5\_0.9'' denotes the using of Eq. (\ref{equ:pbs_split_two}) with $Bound_{best\_pos}=0.4$, $Bound_1=0.4$, $Score_1=0.8$, $Bound_2=0.5$ and $Score_2=0.9$.}\label{fig:Detailed experimental results of different Precise Box Score(PBS)}
\end{figure}

Fig. \ref{fig:Detailed experimental results of different Precise Box Score(PBS)} shows the detailed experimental results of Precise Sigmoid+PBS, and the PBS strategy of ``split\_0.4\_0.8\_0.5\_0.9'' gets the best accuracy. Next, results in Table \ref{table:accuracy comparison of precise sigmoid+PBS and softmax} prove the effectiveness of PBS. The network using Precise Sigmoid+PBS can consistently get accuracy gain compared with conventional softmax+original training strategy.






\begin{table}[]
\centering
\scalebox{0.65}
{
\begin{tabular}{|c|c|c|c|}
\hline
\multirow{2}{*}{Architecture} & \multicolumn{3}{c|}{Accuracy(\%)} \\ \cline{2-4}
                             & 1000 FP    & 500 FP    & 100 FP    \\ \hline  \hline
ZF-net(softmax)&92.5&91.2&83.3 \\ \hline
ZF-net(Precise Sigmoid, no PBS)&92.7&90.7&83.2 \\ \hline  \hline
ZF-24-net(softmax)&90.6&88.9&82.2 \\ \hline
ZF-24-net(Precise Sigmoid, no PBS)&89.8&88.2&82.3 \\ \hline \hline
VGG\_CNN\_M\_1024(softmax)&91.9&90.5&83.6 \\ \hline
VGG\_CNN\_M\_1024(Precise Sigmoid, no PBS)&92.2&90.5&82.3 \\ \hline \hline
VGG16(softmax)&94.1&92.8&86.2 \\ \hline
VGG16(Precise Sigmoid, no PBS)&95.0&93.6&82.2 \\ \hline
\end{tabular}
}
\vspace{1em}
\caption{Accuracy(\%) comparison between Precise Sigmoid with softmax on FDDB. Both of the two networks are trained by original R-CNN style training strategy. And the loss weight $\lambda=300$ for $L_{PreciseSigmoid}$}
\label{table:accuracy comparison of precise sigmoid and softmax}
\end{table}

\begin{table}[]
\centering
\scalebox{0.65}
{
\begin{tabular}{|c|c|c|c|}
\hline
\multirow{2}{*}{Architecture} & \multicolumn{3}{c|}{Accuracy(\%)} \\ \cline{2-4}
                             & 1000 FP    & 500 FP    & 100 FP    \\ \hline  \hline
ZF-net(softmax)&92.5&91.2&83.3 \\ \hline
ZF-net(Precise Sigmoid)&92.7&90.7&83.2 \\ \hline
\textbf{ZF-net(Precise Sigmoid+PBS)}&\textbf{94.6}&\textbf{93.3}&\textbf{83.4} \\ \hline \hline
ZF-24-net(softmax)&90.6&88.9&82.2 \\ \hline
ZF-24-net(Precise Sigmoid)&89.8&88.2&82.3 \\ \hline
\textbf{ZF-24-net(Precise Sigmoid+PBS)}&\textbf{91.7}&\textbf{90.2}&\textbf{83.1} \\ \hline \hline
VGG\_CNN\_M\_1024(softmax)&91.9&90.5&83.6 \\ \hline
VGG\_CNN\_M\_1024(Precise Sigmoid)&92.2&90.5&82.3 \\ \hline
\textbf{VGG\_CNN\_M\_1024(Precise Sigmoid+PBS)}&\textbf{93.7}&\textbf{92.3}&\textbf{81.8} \\ \hline \hline
VGG16(softmax)&94.1&92.8&86.2 \\ \hline
VGG16(Precise Sigmoid)&95.0&93.6&82.2 \\ \hline
\textbf{VGG16(Precise Sigmoid+PBS)}&\textbf{95.4}&\textbf{94.5}&\textbf{87.4} \\ \hline
\end{tabular}
}
\vspace{1em}
\caption{Accuracy(\%) comparison of Precise Sigmoid+PBS and softmax.}
\label{table:accuracy comparison of precise sigmoid+PBS and softmax}
\end{table}

\subsection{Precise Sigmoid+PBS on WIDER FACE}\label{subsec:Precise Sigmoid+PBS on WIDER face}

We also evaluate the performance of Precise Sigmoid+PBS on WIDER FACE. SSH~\cite{najibi2017ssh} is used as baseline, using the train set of WIDER FACE for training. The training and testing are both single scale, meaning we rescale the shorter side of the image up to 1200 pixels while keeping the longer side below 1600 pixels without changing the aspect ratio. The settings of scale and aspect ratio follow the SSH's.

Table \ref{table:Comparison of original SSH with SSH trianed by our Precise Sigmoid+PBS} compares the original SSH with SSH trained by our Precise Sigmoid+PBS. SSH uses two rough IoU thresholds, 0.5 and 0.3. In experiments, we adjust the threshold for positive samples on the architecture of Precise Sigmoid. We also use the empirical best parameters of PBS(``pos0.4+split\_0.4\_0.8\_0.5\_0.9'', the results of Fig \ref{fig:Detailed experimental results of different Precise Box Score(PBS)}), to demonstrate the effectiveness of Precise Sigmoid+PBS. Table \ref{table:Comparison of original SSH with SSH trianed by our Precise Sigmoid+PBS} shows the effectiveness of Precise Sigmoid+PBS:

(a) SSH trained by Precise Sigmoid+PBS outperforms the original SSH~\cite{najibi2017ssh} by 0.3\%, 0.6\% and 0.8\% in ``easy'', ``medium'', ``hard'' subsets of WIDER FACE respectively.

(b) current result of SSH trained by Precise Sigmoid+PBS is just simply using the setting of ``pos0.4+split\_0.4\_0.8\_0.5\_0.9'', which is the empirical parameters of PBS, as shown in Fig. \ref{fig:Detailed experimental results of different Precise Box Score(PBS)}. Specific adjustment of parameters of PBS may make the model get even better result.

\begin{table}[]
\centering
\scalebox{0.65}
{
\begin{tabular}{|c|c|c|c|}
\hline
\multirow{2}{*}{Method} & \multicolumn{3}{c|}{Accuracy(\%)} \\ \cline{2-4}
                             & easy    & medium    & hard    \\ \hline  \hline

SSH(softmax)~\cite{najibi2017ssh}(pos0.5)&91.9&90.7&81.4 \\ \hline
SSH(Precise Sigmoid)(pos0.4)&92.3&90.5&79.0 \\ \hline
SSH(Precise Sigmoid)(pos0.5)&91.8&90.5&81.3 \\ \hline
SSH(Precise Sigmoid)(pos0.7)&88.3&85.8&74.0 \\ \hline
\textbf{SSH(Precise Sigmoid+PBS)(pos0.4+split\_0.4\_0.8\_0.5\_0.9)}&\textbf{92.2}&\textbf{91.3}&\textbf{82.2} \\ \hline
\end{tabular}
}
\vspace{1em}
\caption{Comparison of original SSH with SSH trained by our Precise Sigmoid+PBS on WIDER FACE. ``pos0.5'' denotes the using of Eq. (\ref{equ:pbs_just_adjust_pos_bound}) with $Bound_{pos}=0.5$. ``pos0.4+split\_0.4\_0.8\_0.5\_0.9'' denotes the using of Eq. (\ref{equ:pbs_split_two}) with $Bound_{best\_pos}=0.4$, $Bound_1=0.4$, $Score_1=0.8$, $Bound_2=0.5$ and $Score_2=0.9$. By default, we set the anchors with IoUs lower than 0.3 as negative samples.}
\label{table:Comparison of original SSH with SSH trianed by our Precise Sigmoid+PBS}
\end{table}

\subsection{SEMCM on FDDB}

There, we use the one stage face detector with main bone of ZF-net as teacher model. And the one stage face detector with main bone of ZF-24-net as student model. As illustrated in Fig. \ref{fig:overall architecture of SEMCM}, we conduct the SEMCM as described in Section \ref{sec:A Simple but Effective Model Compression Method(SEMCM)}. The teacher model(ZF-net) has accuracy of 94.6, 93.3 and 83.4 when the false positive on FDDB is 1000, 500 and 100, and the model is trained by Precise Sigmoid+PBS. The student model(ZF-24-net) is all layers' channel reduced by $\frac{1}{4}$ from teacher model(ZF-net), except the layers for classification and regression. The student model(ZF-24-net) is half-trained with accuracy of 88.4, 86.5 and 77.4.

\begin{figure}
\centering
\includegraphics[width=0.4\textwidth]{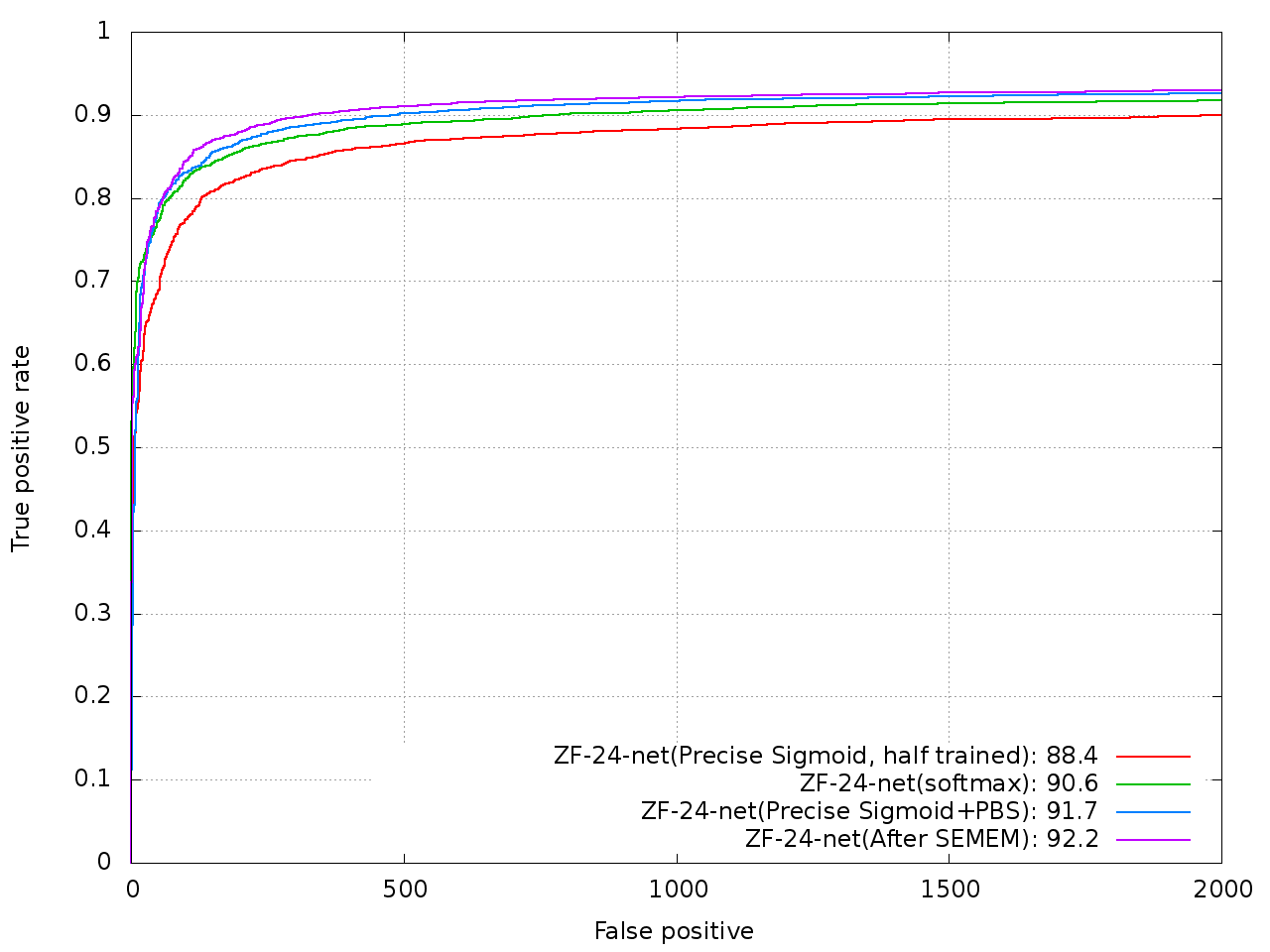}
\caption{Results of SEMCM.}\label{fig:Results of SEMCM}
\end{figure}

Table \ref{table:Results of SEMCM} and Fig \ref{fig:Results of SEMCM} shows that: after using SEMCM, the performance of small student model can be raised to 92.2, 91.0, 84.5, even better than the accuracy of the small student model trained with Precise Sigmoid+PBS. The detailed accuracy gain is(the accuracy shown next is when false positive on FDDB is 1000, 500 and 100 respectively.):

(a) The small model has model size of 1M. And it is trained from the performance of 88.4, 86.5 and 77.4, which is a half-trained model.

(b) The small model is raised to the accuracy of 92.2, 91.0, 84.5 from 88.4, 86.5 and 77.4. The gain is 3.8, 4.5, 7.1, relative to the half-trained model for initialization. The gain is 1.6, 2.1, 2.3, relative to the model trained with softmax. The gain is 0.5, 0.8, 1.4, relative to the model trained with Precise Sigmoid+PBS.

(c) The small model is raised to the accuracy of 92.2, 91.0, 84.5, even better than the small model trained with Precise Sigmoid+PBS, proving the complementary of Precise Sigmoid+PBS and SEMCM.

\begin{table}[]
\centering
\scalebox{0.45}
{
\begin{tabular}{|c|c|c|c|}
\hline
\multirow{2}{*}{Architecture} & \multicolumn{3}{c|}{Accuracy(\%)} \\ \cline{2-4}
                             & 1000 FP    & 500 FP    & 100 FP    \\ \hline  \hline
ZF-24-net(Precise Sigmoid, half trained)(For student model initialization)&88.4&86.5&77.4 \\ \hline
ZF-24-net(softmax)&90.6&88.9&82.2 \\ \hline
ZF-24-net(Precise Sigmoid+PBS)&91.7&90.2&83.1 \\ \hline
\textbf{ZF-24-net(\textbf{After SEMCM})}&\textbf{92.2}&\textbf{91.0}&\textbf{84.5} \\ \hline
\end{tabular}
}
\vspace{1em}
\caption{Results of SEMCM....}
\label{table:Results of SEMCM}
\end{table}

\subsection{Qualitative Results}

\begin{figure*}
\centering
\subfigure[Qualitative results of one stage face detector of ZF-24-net as main bone(Precise Sigmoid+PBS+SEMEM)]
{
\includegraphics[width=1\textwidth]{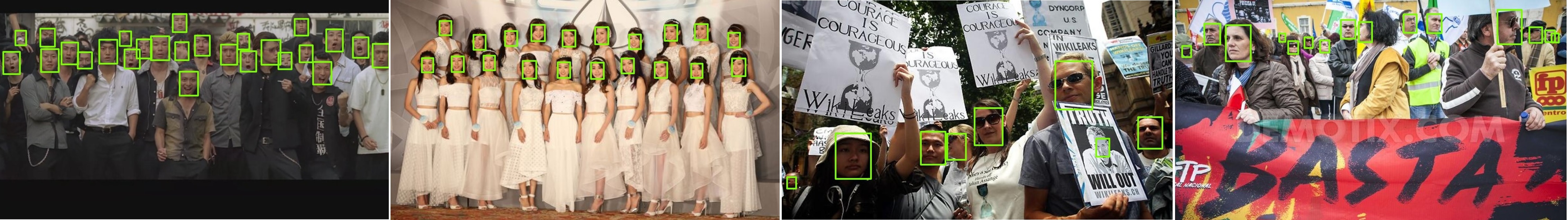}
}
\hspace{0in}
\subfigure[Qualitative results of one stage face detector of ZF-net as main bone(Precise Sigmoid+PBS)]
{
\includegraphics[width=1\textwidth]{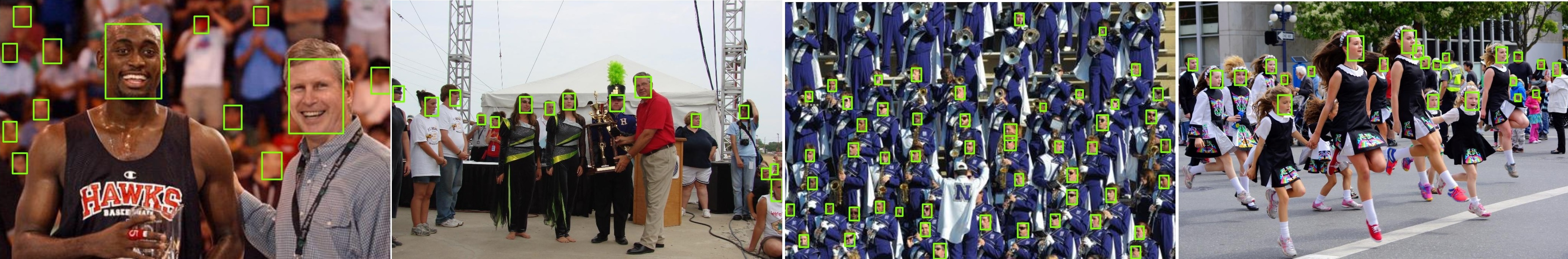}
}
\hspace{0in}
\subfigure[Qualitative results of SSH trained by Precise Sigmoid+PBS]
{
\includegraphics[width=1\textwidth]{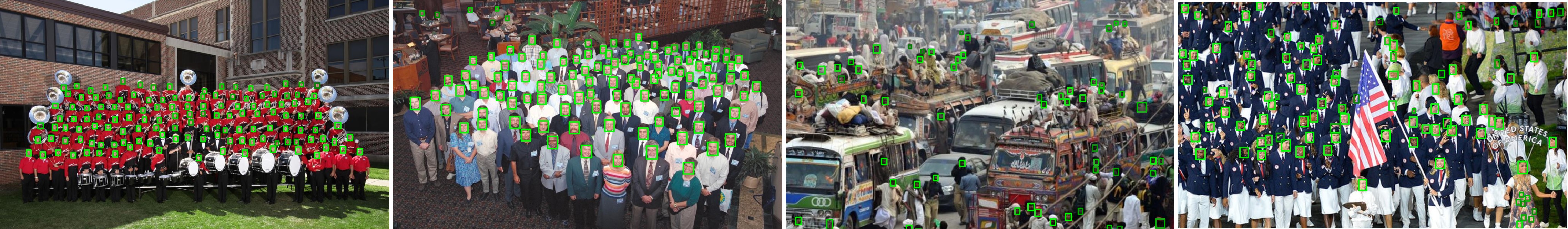}
}
\caption{Qualitative results}\label{fig:Qualitative results}
\end{figure*}

Some qualitative results of three face detectors are shown in Fig \ref{fig:Qualitative results}. The three models have model size of 1.1M, 17.3M and 98.1M, respectively. The first one stage face detector of ZF-24-net as main bone is trained by Precise Sigmoid+PBS+SEMEM. The second one stage face detector of ZF-net as main bone is trained by Precise Sigmoid+PBS. And the third model, SSH, is trained by Precise Sigmoid+PBS.

\section{Conclusion}\label{sec:Conclusion}

We propose a novel training strategy, Precise Box Score(PBS), which can extract more information from detection dataset and benefit the post-processing of NMS for the precise bounding box scores. And a new architecture, Precise Sigmoid, is introduced for the implementation of PBS. We do experiments using one stage face detector on FDDB to explore how to design the function of PBS. Further more, a simply but effective model compression method(SEMCM) is proposed for one stage face detector, which can boost the performance of face detection further. Experiments demonstrate: (a) Precise Sigmoid+PBS can consistently improve the performance of face detection, and (b) the complementary of Precise Sigmoid+PBS and SEMCM.

{\small
\bibliographystyle{ieee}
\bibliography{egbib}
}

\end{document}